\documentclass[conference]{IEEEtran}
\IEEEoverridecommandlockouts
\usepackage{cite}
\usepackage{amsmath,amssymb,amsfonts}
\usepackage{algorithmic}
\usepackage{graphicx}
\usepackage{textcomp}
\usepackage{xcolor}
\usepackage{booktabs}
\usepackage{siunitx}
\usepackage{subcaption}
\usepackage[colorlinks=true,
            linkcolor=blue,      % internal links (e.g., \ref{...}, \pageref{...})
            citecolor=green,     % citations (e.g., \cite{...})
            urlcolor=red         % URLs (e.g., \url{...})
           ]{hyperref}

\usepackage[numbers,sort&compress]{natbib}
\usepackage{cite}

\def\BibTeX{{\rm B\kern-.05em{\sc i\kern-.025em b}\kern-.08em
    T\kern-.1667em\lower.7ex\hbox{E}\kern-.125emX}}
\begin{document}

\title{Cybersecurity-Focused Anomaly Detection in Connected Autonomous Vehicles Using Machine Learning}

\author{
\small 
% 第一行
\begin{minipage}[t]{0.3\textwidth}
  \centering
  \textbf{Prathyush Kumar Reddy Lebaku}\\
  Engineering Data Science\\
  University of Houston\\
  Houston, TX, USA\\
  \texttt{plebaku@cougarnet.uh.edu}
\end{minipage}
\hfill
\begin{minipage}[t]{0.3\textwidth}
  \centering
  \textbf{Lu Gao, Ph.D.}\\
  Department of Civil and Environmental Engineering\\
  University of Houston\\
  Houston, TX, USA\\
  \texttt{lgao5@central.uh.edu}
\end{minipage}
\hfill
\begin{minipage}[t]{0.3\textwidth}
  \centering
  \textbf{Yunpeng Zhang, Ph.D.}\\
  Department of Information Science Technology\\
  University of Houston\\
  Houston, TX, USA\\
  \texttt{yzhang119@uh.edu}
\end{minipage}
\\[1em]  % 强制换行
\hfill
\begin{minipage}[t]{0.3\textwidth}
  \centering
  \textbf{Zhixia Li, Ph.D.}\\
  Department of Civil and Architectural Engineering and Construction Management\\
  University of Cincinnati\\
  Cincinnati, OH, USA\\
  \texttt{lizx@ucmail.uc.edu}
\end{minipage}
\hfill
\begin{minipage}[t]{0.3\textwidth}
  \centering
  \textbf{Yongxin Liu, Ph.D.}\\
  Mathematics Department\\
  Embry-Riddle Aeronautical University\\
  Daytona Beach, FL, USA\\
  \texttt{liuy11@erau.edu}
\end{minipage}
% 第二行
\begin{minipage}[t]{0.3\textwidth}
  \centering
  \textbf{Tanvir Arafin, Ph.D.}\\
  Department of Cyber Security Engineering\\
  George Mason University\\
  Fairfax, VA, USA\\
  \texttt{marafin@gmu.edu}
\end{minipage}
}

% \author{%
% Prathyush Kumar Reddy Lebaku\textsuperscript{1a},
% Lu Gao, Ph.D.\textsuperscript{1a},
% Yunpeng Zhang, Ph.D.\textsuperscript{1a},\\  % ← 在合适位置断行
% Zhixia Li, Ph.D.\textsuperscript{2b},
% Yongxin Liu, Ph.D.\textsuperscript{3c},
% Tanvir Arafin, Ph.D.\textsuperscript{4d}
% \\[0.8ex]
% \textsuperscript{1}Department of Civil and Environmental Engineering, University of Houston, Houston, TX, USA\quad \\
% \textsuperscript{2}University of Cincinnati, Cincinnati, OH, USA\quad \\
% \textsuperscript{3}Embry-Riddle Aeronautical University, Daytona Beach, FL, USA\quad \\
% \textsuperscript{4}George Mason University, Fairfax, VA, USA\\[0.6ex]
% \textsuperscript{a}\{plebaku, lgao5, yzhang119\}@uh.edu,\;
% \textsuperscript{b}lizx@ucmail.uc.edu,\;
% \textsuperscript{c}liuy11@erau.edu,\;
% \textsuperscript{d}marafin@gmu.edu%
% }

\maketitle

\begin{abstract}

Anomaly detection in connected autonomous vehicles (CAVs) is crucial for maintaining safe and reliable transportation networks, as CAVs can be susceptible to sensor malfunctions, cyber-attacks, and unexpected environmental disruptions. This study explores an anomaly detection approach by simulating vehicle behavior, generating a dataset that represents typical and atypical vehicular interactions. The dataset includes time-series data of position, speed, and acceleration for multiple connected autonomous vehicles. We utilized machine learning models to effectively identify abnormal driving patterns. First, we applied a stacked Long Short-Term Memory (LSTM) model to capture temporal dependencies and sequence-based anomalies. The stacked LSTM model processed the sequential data to learn standard driving behaviors. Additionally, we deployed a Random Forest model to support anomaly detection by offering ensemble-based predictions, which enhanced model interpretability and performance. The Random Forest model achieved an R² of 0.9830, MAE of 5.746, and a 95th percentile anomaly threshold of 14.18, while the stacked LSTM model attained an R² of 0.9998, MAE of 82.425, and a 95th percentile anomaly threshold of 265.63. These results demonstrate the models’ effectiveness in accurately predicting vehicle trajectories and detecting anomalies in autonomous driving scenarios.

\end{abstract}

\begin{IEEEkeywords}
Cybersecurity, Connected Autonomous Vehicles (CAV), Deep Learning, Machine Learning, Anomaly Detection
\end{IEEEkeywords}

\section{Introduction}

Connected Autonomous Vehicles (CAVs) have gained attention for improving road safety, reducing traffic, and increasing transportation efficiency \citep{guler2014using, siegel2017survey, doecke2015real, beak2017adaptive, tideman2013simulation, ghoul2021real}. However, their integration into real-world traffic highlights the need for robust anomaly detection systems. These systems are essential for identifying issues in real-time, ensuring safe operation, and preventing accidents or failures. While CAVs offer many advantages, they are also vulnerable to cyber-attacks that can compromise safety and functionality \citep{salek2022review,gao2025exploring,rajbahadur2018survey}. The following review summarizes recent advancements in anomaly detection for CAVs, focusing on datasets, cyber-attack scenarios, general applications, and the models used.

\subsection{Real-World vs. Simulated Datasets}
Real-world datasets have been used in anomaly detection modeling, such as the Safety Pilot Model Deployment (SPMD) dataset \citep{eziama2020detection, wang2020anomaly,khanmohammadi2024time}, the Car Hacking Dataset \citep{mansourian2023deep}, and the Federal Highway Administration (FHWA) Platooning Dataset \citep{wang2024anomaly}. The SPMD dataset was created by the U.S. Department of Transportation. It includes features such as Basic Safety Messages (BSMs), vehicle trajectories, speed, acceleration, position, heading, timestamps, and communication event data from Vehicle-to-Vehicle (V2V) and Vehicle-to-Infrastructure (V2I) interactions. The dataset primarily involves human-driven vehicles, with a mix of manual and semi-automated systems. Approximately 2,800 vehicles participated in the deployment \citep{gay2015safety}. The Car Hacking Dataset was created by researchers at Korea University to study how cars can be attacked by hackers and how to defend against those attacks. They used a real car driven by a human and recorded the messages that the car's internal computer systems send to each other. These messages control things like the engine, brakes, and steering \citep{kang2021car}. The Federal Highway Platoon Dataset was created by the Federal Highway Administration (FHWA) to study how Connected and Automated Vehicles (CAVs) and human-driven vehicles interact in traffic. The dataset includes information like vehicle speed, position, and interactions on freeways and arterials. It involves both human-driven vehicles and CAVs, with multiple vehicles included in the experiments. The goal was to analyze driving behavior and improve traffic models \citep{tiernan2017test}.

Simulated datasets are also widely used to test anomaly detection methods in controlled environments. For example, \citet{wang2019location} generated a dataset using OMNeT++ and SUMO to simulate a 2 km × 2 km area of Bristol, UK, with 150 vehicles exchanging normal and falsified location data, modeling anomalies like falsified positions based on transmitter and receiver locations and RSSI values. \citet{devika2024vadgan} utilized the VeReMi Extension dataset, which included vehicle dynamics like position, speed, acceleration, and heading under simulated spoofing and Sybil attacks, generated using OMNeT++, Simulation of Urban MObility (SUMO) and Luxembourg SUMO Traffic data. \citet{dong2020impact} employed a dataset based on Cooperative Adaptive Cruise Control (CACC) models to study the effects of cyber-attacks on traffic, focusing on vehicle parameters like speed, acceleration, and position. \citet{wang2021deep} used a SUMO-generated dataset to simulate vehicle-following scenarios with time-series data on positions, velocities, accelerations, and relative distances of ego and leading vehicles, modeling stealthy attacks like shifting, scaling, and time-variant manipulation to evaluate ACC systems.

\subsection{Model used for Anomaly Detection}

Early research used statistical and physical models to detect anomalies in real-world systems. For example, \citet{wang2020anomaly} developed the Augmented Extended Kalman Filter (AEKF), an improved version of the Kalman Filter that handled biases and communication delays. It used a $\chi^2$ fault detector to classify anomalies, making it effective in dynamic traffic environments. \citet{dong2020impact} applied a Cooperative Adaptive Cruise Control (CACC) car-following model with Lyapunov stability analysis to study cyber-attacks on traffic flow. This approach detected anomalies by analyzing deviations in stability and congestion, highlighting system vulnerabilities. These works emphasized the importance of real-time detection.

Deep learning revolutionized anomaly detection by modeling complex temporal patterns. \citet{malhotra2015long} introduced stacked LSTM networks for time-series anomaly detection, using reconstruction errors to capture short- and long-term dependencies. \citet{wang2019location} extended this with Deep Autoencoders to detect positional anomalies in vehicle data, effectively handling large deviations but struggling with subtle anomalies. To address this, \citet{wang2024anomaly} proposed LAGMM, combining LSTMs for temporal modeling with Gaussian Mixture Models (GMM) for density-based anomaly detection, achieving robust performance in real-world scenarios. Hybrid models also emerged, blending multiple techniques for better results. \citet{eziama2020detection} developed DWT-BDL, integrating Discrete Wavelet Transform for denoising with Bayesian Deep Learning for probabilistic detection, excelling in noisy environments. \citet{khanmohammadi2024time} introduced D-CNN-LSTM Autoencoder and CNN-BiLSTM models, combining CNNs for spatial features with LSTMs for temporal data, effectively detecting subtle anomalies. \citep{mansourian2023deep} advanced this with an LSTM-ConvLSTM model and Gaussian Naive Bayes, capturing spatiotemporal patterns in CAN bus data and detecting attacks like fuzzy and spoofing efficiently. Generative models also brought advancements. \citet{devika2024vadgan} proposed VADGAN, an LSTM-based GAN optimized with Mean Squared Error and log-cosh loss functions, reconstructing sequential data to detect subtle anomalies like GPS spoofing and replay attacks, offering a versatile approach for diverse scenarios.

\subsection{Types of Cyber Attacks}
There are three different types of attacks mentioned by the papers which we reviewed. The first one is internal attack which focuses on manipulating or exploiting vulnerabilities within the vehicles onboard systems and networks, such as CAN bus and GPS and other vehicle sensor systems \citep{sun2021survey,eziama2020detection,mansourian2023deep,khanmohammadi2024time}. And the second is external communication attack which targets communication between vehicles (V2V) or between V2I. The common type of attacks included in this are DoS Attack (the attacker floods the V2V or V2I communication channels with and overwhelming amount of data which causes communication failure) and message falsification and spoofing \citep{wang2020anomaly,eziama2020detection,wang2021deep,devika2024vadgan,wang2024anomaly} And the third one is position and speed manipulation attack which focuses on modifying speed (false speed data into vehicle control system, this causes the vehicle to regulate the speed incorrectly) and attack position (inject falsified position data into the vehicle GPS system or V2I/V2V Communication channel) data \citep{wang2019location,dong2020impact}. 

Previous studies did not examine cases where a vehicle sends false location data to nearby vehicles. In our research, we studied a cyber-attack where a vehicle is tricked into thinking another vehicle is farther away than it actually is. This false perception disrupts safe driving, leading to sudden acceleration or braking, and reduces overall traffic safety. By analyzing this type of attack using location data in a network of vehicles, our work aims to improve detection methods and protect connected vehicles from complex cyber threats.

\section{Methodology}

\begin{figure}[htbp]
    \centering
    \includegraphics[width=1\linewidth]{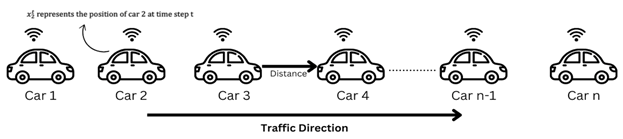}
    \caption{Car Following Trajectory}
    \label{fig:overview}
\end{figure}

In this research, we aim to tackle the challenge of anomaly detection in CAV environments, particularly when these vehicles face cybersecurity threats. The focus is on developing a robust system that can detect anomalies using limited sensor inputs, specifically vehicle position data. Figure \ref{fig:overview} depicts a scenario where n connected autonomous vehicles are traveling along a single-lane road. These vehicles form a wireless network, allowing them to continuously exchange information about their speed and positions. This communication enables each vehicle to make dynamic adjustments, such as modifying speed or maintaining safe following distances. By analyzing the shared position data, our approach seeks to identify anomalies that may signal cyber-attacks or system malfunctions, ensuring robust, real-time detection without the need for extensive multi-sensor inputs.

After establishing a connected environment where n vehicles share positional data over a wireless network, the next step is to develop models capable of detecting anomalies within this framework. To achieve this, we first focus on leveraging machine learning techniques to predict vehicle behavior based on historical position data. Given the continuous exchange of information among vehicles, we hypothesize that deviations from predicted trajectories could indicate potential anomalies, such as cyber-attacks or system malfunctions. For this purpose, we begin with the Random Forest (RF) model, which is well-suited for handling non-linear relationships and capturing patterns in time-series data.

\subsection{Random Forest Model}

The Random Forest (RF) model was chosen as our initial approach due to its robustness and ability to handle high-dimensional data. In previous studies within the transportation domain, RF has demonstrated consistently reliable performance across a variety of applications \citep{cheng2019applying,manzoor2018vehicle,gao2021deep,parosh2025united,elsadig2025connected,gao2022missing}. In this case, the RF model uses the positions from the first $T_1$ time steps and predicts next $T_2$ time steps. The input X can be expressed as

\begin{equation}
    \textbf{X} = (\textbf{X}_1, \textbf{X}_2, \ldots, \textbf{X}_T)^\prime
\end{equation}

\noindent where $T$ is the total number of data points (number of time steps during the simulation) and $\textbf{X}_t$ represent the $t^{th}$ data point, which can be represented as

\begin{equation}  
       \textbf{X}_t=(\textbf{x}^t,\textbf{x}^{t+1},…,\textbf{x}^{t+T_1-1}), t=1,2,…,T
\end{equation}

\noindent where $\textbf{x}^t=\{x_1^t,x_2^t,…,x_n^t\}$ and $x_i^t$ represents the location of the $i^{th}$ vehicle at time \(t\) (the distance between the starting point and the current location of the $i^{th}$ vehicle). 

The target $Y$ can be expressed as

\begin{equation}
\textbf{Y}=(\textbf{Y}_1,\textbf{Y}_2,…,\textbf{Y}_T)'    
\end{equation}

\noindent where,

\begin{equation}\label{eq:eq4}
       \textbf{Y}_t=(\textbf{x}^{T_1+t},\textbf{x}^{T_1+t+1},…,\textbf{x}^{T_1+t+T_2-1}),t=1,2,…,T  
\end{equation}

\noindent where $\textbf{x}^t={x_1^t,x_2^t,…,x_n^t }$. As shown in Equation \eqref{eq:eq4}, for each data point, the target is multiple output. To detect anomalies, we calculate the prediction error by comparing the actual position $Y_t$ with the predicted position $\hat{\textbf{Y}}_t$.

\begin{equation}
    \hat{\textbf{Y}}_t  = (\hat{\textbf{x}}^{T_1+t},\hat{\textbf{x}}^{T_1+t+1},...,\hat{\textbf{x}}^{T_1+t+T_2-1)},t=1,2,…,T
\end{equation}

\noindent where,\\
$\textbf{Y}_t$ = the actual positions of the vehicles for the next $T_2$ time steps;\\
$\hat{\textbf{Y}}_t$ = the predicted positions for the same time steps using your model.

The difference between the actual and predicted is calculated as:
\begin{equation}
	\textbf{D}_t  = |\textbf{Y}_t-\hat{\textbf{Y}}_t|	
\end{equation}

\noindent where each element of $\textbf{D}_t$ is defined as,

\begin{equation}
	d_i=x_i^{T_1+t-1}-\hat{x}_i^{T_1+t-1}, i=1,…,n; t=1,…,T_2
\end{equation}

\noindent where,\\
$x_i^{T_1+t-1}$ = the actual position of vehicle $i$ at time step $t$;\\
$\hat{x}_i^{T_1+t-1}$ = the predicted position of vehicle $i$ at time step $t$;\\
$d_i$ = the difference in position for the $i$th vehicle at time step $t$. \\

After computing the difference matrix $\textbf{D}_t$ , we computed the sum of the absolute differences to quantify the total prediction error: 
\begin{equation}  
	S=\sum_{i=1}^n \sum_{t=1}^{T_2} |d_i|
\end{equation}

where $S$ represents the total prediction error over all vehicles and future time steps. To detect anomalies, we compare S against a predefined threshold:
\begin{equation}
	  S > \theta
\end{equation}      

If $S$ exceeds this  $\theta$ , it indicates that the predicted trajectory deviates significantly from the actual trajectory, and the trajectory is flagged as anomalous. The threshold is determined through data modeling and analysis of normal trajectory data.

\subsection{Stacked LSTM Model}
\begin{figure}[htbp]
    \centering
    \includegraphics[width=1\linewidth]{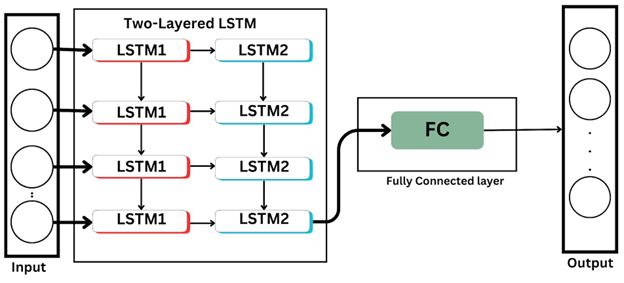}
    \caption{Stacked LSTM Architecture}
    \label{fig:lstm}
\end{figure}

We also tested the stacked LSTM architecture model, which is specifically designed to forecast future vehicle positions by leveraging historical trajectory data (Figure \ref{fig:lstm}). The model uses a sequence of past time steps as input to predict the subsequent time steps. This input data consists of the position information of multiple vehicles, with each vehicle's trajectory represented solely by its position at each time step. By training on normal, non-anomalous trajectories, the model learns typical motion patterns, enabling it to identify expected behaviors within the dataset. The inclusion of multiple time steps in both the input and output allows the model to effectively capture temporal dependencies and patterns across the vehicle trajectories, enhancing its predictive accuracy. Prior work has demonstrated the effectiveness of LSTM models in time-series forecasting tasks \cite{siami2019performance,kumar2021lstm,gao2023deep,wang2021inclstm}.

% --- Sliding-window input to the LSTM ----------------------------------------
The input data to the LSTM model are arranged as a sliding window of past
positions over \(T_{1}\) time steps, covering the positions of \(n\) cars.  
For each time step \(t\), the window is

\begin{equation}
\mathbf{X}
  = \bigl(\mathbf{X}_{1},\,\mathbf{X}_{t+1},\,\dots,\,\mathbf{X}_{t+T_{1}-1}\bigr)',
\tag{10}
\end{equation}

and
\begin{equation}
\mathbf{X}_{t}
  = \bigl(\mathbf{x}^{t},\,\mathbf{x}^{t+1},\,\dots,\,\mathbf{x}^{t+T_{1}-1}\bigr)^{\prime},
  \qquad t = 1,2,\dots,T .
\end{equation}

A single data point \(\mathbf{X}\) is therefore a \(T \times n\) matrix, where
\(T\) is the number of time steps and \(n\) is the number of cars.  Each  
\(\mathbf{x}^{t} = \{x_{1}^{t},x_{2}^{t},\dots,x_{n}^{t}\}\) contains the positions of all
cars at time step \(t\) (\(t = 1,2,\dots,T\)).

% --- Output sequence ---------------------------------------------------------
\begin{equation}
\mathbf{Y}
  = \bigl(\mathbf{Y}_{1},\,\mathbf{Y}_{2},\,\dots,\,\mathbf{Y}_{T}\bigr)^{\prime},
\tag{11}
\end{equation}
where
\begin{equation}
\mathbf{Y}_{t}
  = \bigl(x^{T_{1}+t},\,x^{T_{1}+t+1},\,\dots,\,x^{T_{1}+t+T_{2}-1}\bigr),
  \qquad t = 1,2,\dots,T .
\tag{12}
\end{equation}

% --- Model prediction --------------------------------------------------------
The model’s predictions are

\begin{equation}
\hat{\mathbf{Y}}_{t}
  = \bigl(\hat{\mathbf{x}}^{T_{1}+t},\,\hat{\mathbf{x}}^{T_{1}+t+1},\,\dots,
          \hat{\mathbf{x}}^{T_{1}+t+T_{2}-1}\bigr),
  \qquad t = 1,2,\dots,T .
\tag{13}
\end{equation}

% --- Error computation -------------------------------------------------------
The difference between actual and predicted positions is

\begin{equation}
\mathbf{D}_{t}
  = \bigl|\mathbf{Y}_{t} - \hat{\mathbf{Y}}_{t}\bigr|,
\tag{14}
\end{equation}
whose elements are

\begin{equation}
d_{i}
  = x_{i}^{T_{1}+t-1} - \hat{x}_{i}^{T_{1}+t-1},
  \qquad i = 1,\dots,n;\; t = 1,\dots,T_{2}.
\tag{15}
\end{equation}

Here, \(x_{i}^{T_{1}+t-1}\) denotes the actual position of vehicle \(i\) at
time step \(t\), and \(\hat{x}_{i}^{T_{1}+t-1}\) is its predicted position. \(d_i\) represents the difference in position for the \(i\)th vehicle at the \(j\)th future time step.  
After computing the difference matrix \(\mathbf{D}_t\), we calculate the sum of absolute differences to quantify the total prediction error:

\begin{equation}
S = \sum_{i=1}^{n} \sum_{t=1}^{T_2} \left| d_i \right|,
\tag{16}
\end{equation}
where \(S\) denotes the total prediction error over all vehicles and future time steps.  
To detect anomalies, we compare \(S\) against a predefined threshold \(\theta\):

\begin{equation}
S > \theta
\tag{17}
\end{equation}
If \(S\) exceeds \(\theta\), it indicates that the predicted trajectory deviates significantly from the actual trajectory, and the trajectory is flagged as anomalous.

\section{Case Study}
Building upon the methodology and models described earlier, this case study demonstrates the effectiveness of our approach in detecting anomalies within a simulated Connected Autonomous Vehicle (CAV) environment. Utilizing the Generalized Intelligent Driver Model (IDM) and the Random Forest and stacked LSTM models, we assess the system’s ability to identify deviations in vehicle trajectories caused by cyber-attacks. This case study simulates autonomous vehicles on a single-lane road communicating via a wireless network. Using the IDM, vehicles calculate acceleration and maintain safe distances, creating a dataset of normal driving behavior. To test anomaly detection, we introduce cyber-attacks such as shifting, noise injections, and time manipulations. A Random Forest model predicts vehicle positions based on past data, while a stacked LSTM model captures long-term patterns to detect anomalies. By evaluating these models against simulated attacks, the study demonstrates their effectiveness in ensuring the safety and security of autonomous vehicle networks.

\subsection{Generalized Intelligent Driver Model (IDM) }

The Generalized Intelligent Driver Model (IDM) is a rule-based framework designed to simulate realistic car-following behavior in autonomous driving scenarios. It considers factors such as desired speed, minimum safe distance, and braking distance, enabling it to mimic human-like driving decisions. By calculating the position, velocity, and acceleration of each vehicle over time, IDM generates continuous trajectory data, which can then be analyzed to detect deviations that may signal potential anomalies.

The acceleration of vehicle \(n\) at time \(t\) is computed as:

\begin{equation}
\dot{v}_n(t) = a \left(1 - \left(\frac{v_n(t)}{v_0}\right)^{\delta} - \left(\frac{s^*(v_n(t), \Delta v_n(t))}{s_n(t)}\right)^2 \right),
\tag{18}
\end{equation}
where the desired minimum spacing \(s^*\) is defined as:

\begin{equation}
s^*(v_n(t), \Delta v_n(t)) = s_0 + T v_n(t) + \frac{v_n(t)\Delta v_n(t)}{2\sqrt{ab}},
\tag{19}
\end{equation}
and the actual spacing \(s_n(t)\) between vehicle \(n\) and its preceding vehicle \(n-1\) is given by:

\begin{equation}
s_n(t) = x_{n-1}(t) - x_n(t) - l,
\end{equation}
with the relative velocity:

\begin{equation}
\Delta v_n(t) = v_n(t) - v_{n-1}(t).
\end{equation}
Here, the model parameters are defined as:\\
\(v_0\) = desired velocity (free traffic velocity);\\
\(s_0\) = minimum spacing between vehicles;\\
\(a\) = maximum acceleration;\\
\(b\) = maximum comfortable deceleration;\\
\(T\) = minimum desired time gap to the vehicle in front;\\
\(\delta\) = acceleration exponent.

Using this information, the subject vehicle applies the Generalized IDM to calculate its acceleration relative to each leading vehicle, selecting the option that ensures the safest outcome \cite{silwal2024assessing}. This is expressed as:

\begin{equation}
\dot{v}_n(t) = \min\left[ \dot{v}_n^{(n-1)}(t), \dot{v}_n^{(n-2)}(t), \dots, \dot{v}_n^{(n-m)}(t) \right],
\tag{20}
\end{equation}
where \(m\) is the number of leading vehicles considered.

\subsection{Types of Cyberattacks}
We used gap manipulation attack where the perceived gap between a vehicle and the one ahead is intentionally altered. This manipulation makes the gap appear smaller or larger than it is, simulating scenarios where a vehicle incorrectly estimates its distance from the one in front, potentially leading to unsafe following behavior.

There are several types of attacks that can compromise the positional data of connected autonomous vehicles. A Shifting Attack involves altering the vehicle’s position by a fixed offset, potentially causing the vehicle to misjudge its surroundings. A Random Noise Attack introduces fluctuations to the positional data, simulating the effects of sensor noise or uneven terrain, which can disrupt the vehicle's ability to maintain accurate positioning. Lastly, a Time-Shift Attack manipulates the timing of the data by delaying or advancing the position reports, mimicking data reporting delays that can lead to incorrect decision-making in real-time driving scenarios.

\subsection{Data Simulation}

This section describes the process of generating trajectory data using the Generalized Intelligent Driver Model (IDM) and the cyberattacks discussed above. We generated two datasets, one including normal trajectories, which follow standard IDM dynamics, and the other including manipulated or "attacked" trajectories that simulate various types of disturbances or anomalies.

\begin{table}[htbp]
\centering
\caption{Simulation Parameters}
\begin{tabular}{|l|c|}
\hline
\textbf{Parameter} & \textbf{Value} \\
\hline
Number of lanes & 1 \\
Length of road segment & 30{,}000 m \\
Desired speed (\(v_0\)) & 10 m/s \\
Length of vehicles (\(l\)) & 5 m \\
Minimum gap (\(s_0\)) & 2 m \\
Maximum acceleration (\(a\)) & 0.73 m/s\(^2\) \\
Maximum deceleration (\(b\)) & 1.67 m/s\(^2\) \\
Acceleration exponent (\(\delta\)) & 4 \\
Number of cars & 10 \\
\hline
\end{tabular}
\label{tab:sim_params}
\end{table}

Using the parameters outlined in Table \ref{tab:sim_params}, we utilized the Generalized Intelligent Driver Model (IDM) to simulate vehicle dynamics along a single-lane, 30,000-meter road segment. Each vehicle maintained a desired speed of 10 m/s, with safe distances defined by a minimum gap of 2 meters and a maximum acceleration of 0.73 m/s². Vehicles were initialized with a length of 5 meters and followed IDM rules to avoid collisions by adjusting acceleration and deceleration based on the gap to the leading vehicle. This setup provided a dataset of normal trajectories under standard IDM dynamics, as well as manipulated trajectories by introducing disturbances to simulate potential anomalies.

In this simulation, the positions of 10 vehicles are recorded at every second ($\Delta$t) up to 3000 seconds, creating a comprehensive trajectory dataset. Each time step captures updated positions for Car 1 through Car 10, sequentially logging data in a CSV file with columns for each car’s position at each second. This dataset structure enables analysis of vehicle positions over time, allowing us to study deviations that may indicate anomalies or disturbances.

\subsection{Normal Traffic Trajectory}

\begin{figure}[htbp]
    \centering
    \includegraphics[width=1\linewidth]{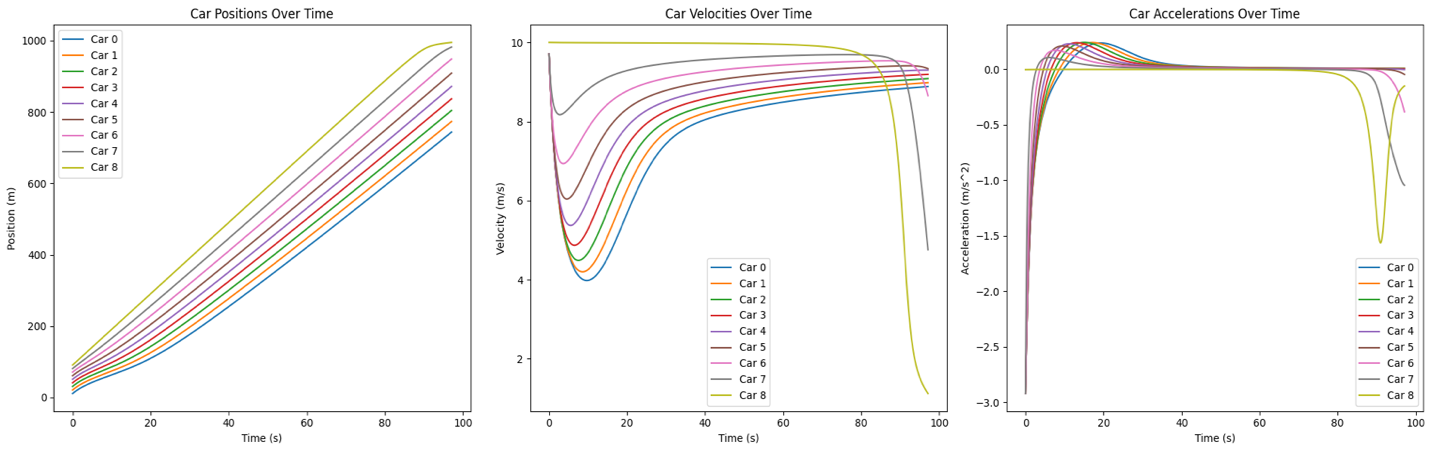}
    \caption{CAV trajectory without any attacks}
    \label{fig:f3}
\end{figure}

Under normal conditions, the vehicles adhered to the IDM rules, producing stable driving behaviors. The simulation recorded the positions, velocities, and accelerations of 10 vehicles every second ($\Delta$t) over a period of 3000 seconds. This continuous recording created a comprehensive trajectory dataset, which was saved in a CSV file format with each row representing the position of Car 1 through Car 10. The results of this simulation are shown in Figure \ref{fig:f3}, where the position plot indicates that vehicles followed smooth, consistent trajectories while maintaining safe distances as they moved along the road. The velocity plot reveals that vehicles quickly accelerated to reach their desired speed of 10 m/s, after which they maintained a steady pace, demonstrating typical driving behavior. The acceleration plot shows minor fluctuations during the initial acceleration phase, with stability achieved as the vehicles reached their target speed. These patterns confirm the stability and regularity of the dataset, providing a reliable baseline against which deviations, potentially caused by cyberattacks or system failures, can be detected.

\subsection{Attacked Traffic Trajectory}
\begin{figure}[htbp]
    \centering
    \includegraphics[width=1\linewidth]{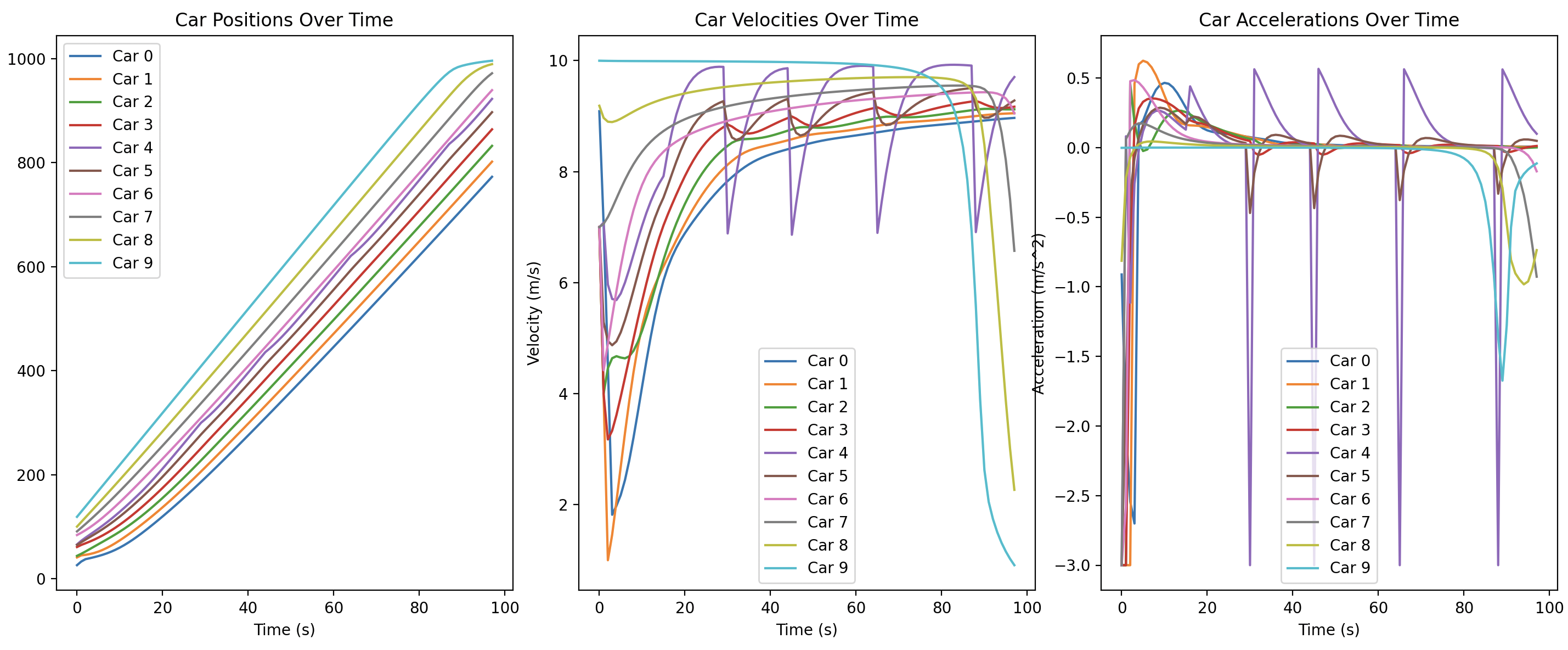}
    \caption{Car 4 perceives the car 5 is far away from its actual distance}
    \label{fig:f4}
\end{figure}

In Figure \ref{fig:f4}, Car 4 perceives Car 5 as being farther away than its actual distance, with a manipulated gap set to 10 times the original gap. This causes Car 4 to oscillate in velocity and acceleration as it compensates for the perceived distance, resulting in instability that disrupts the vehicle’s smooth trajectory.

\begin{figure}[htbp]
    \centering
    \includegraphics[width=1\linewidth]{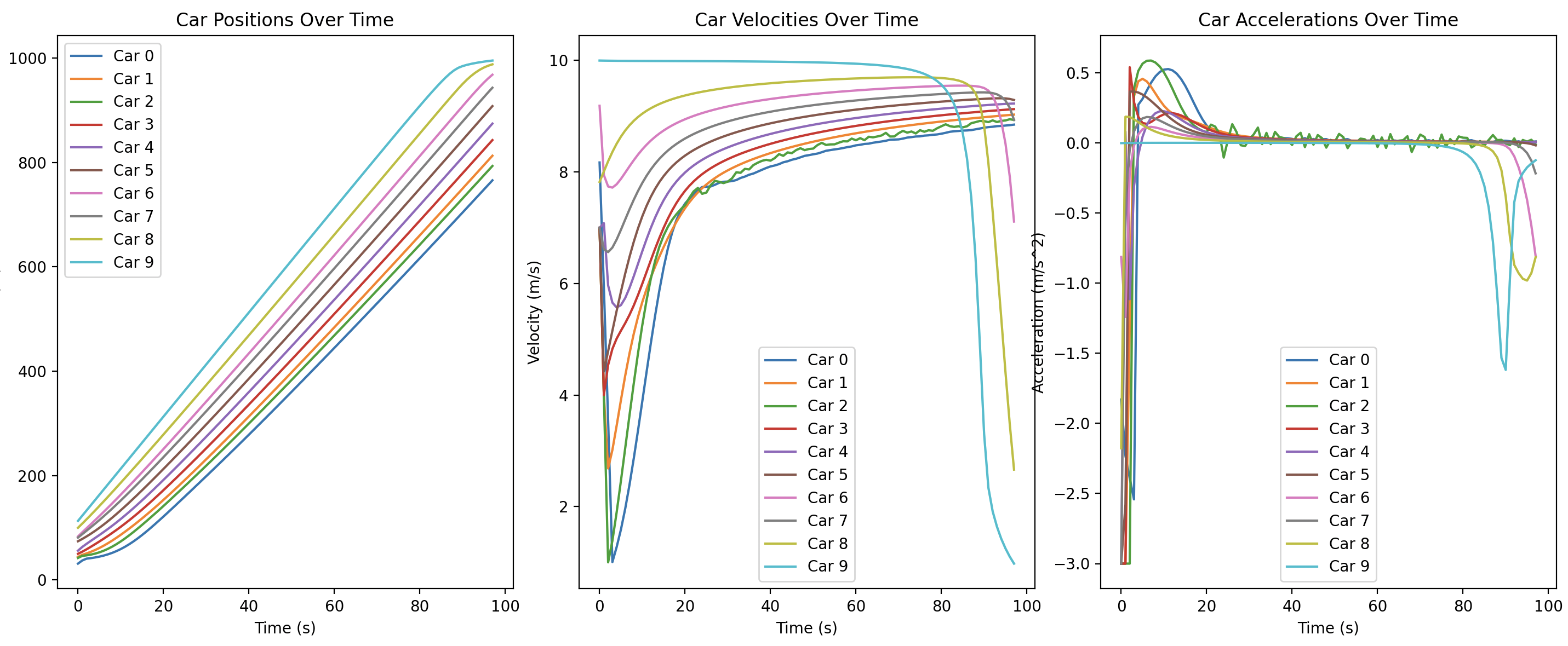}
    \caption{Car 2 experiences a random fluctuation in the perceived gap}
    \label{fig:f5}
\end{figure}

\begin{figure}[htbp]
    \centering
    \includegraphics[width=1\linewidth]{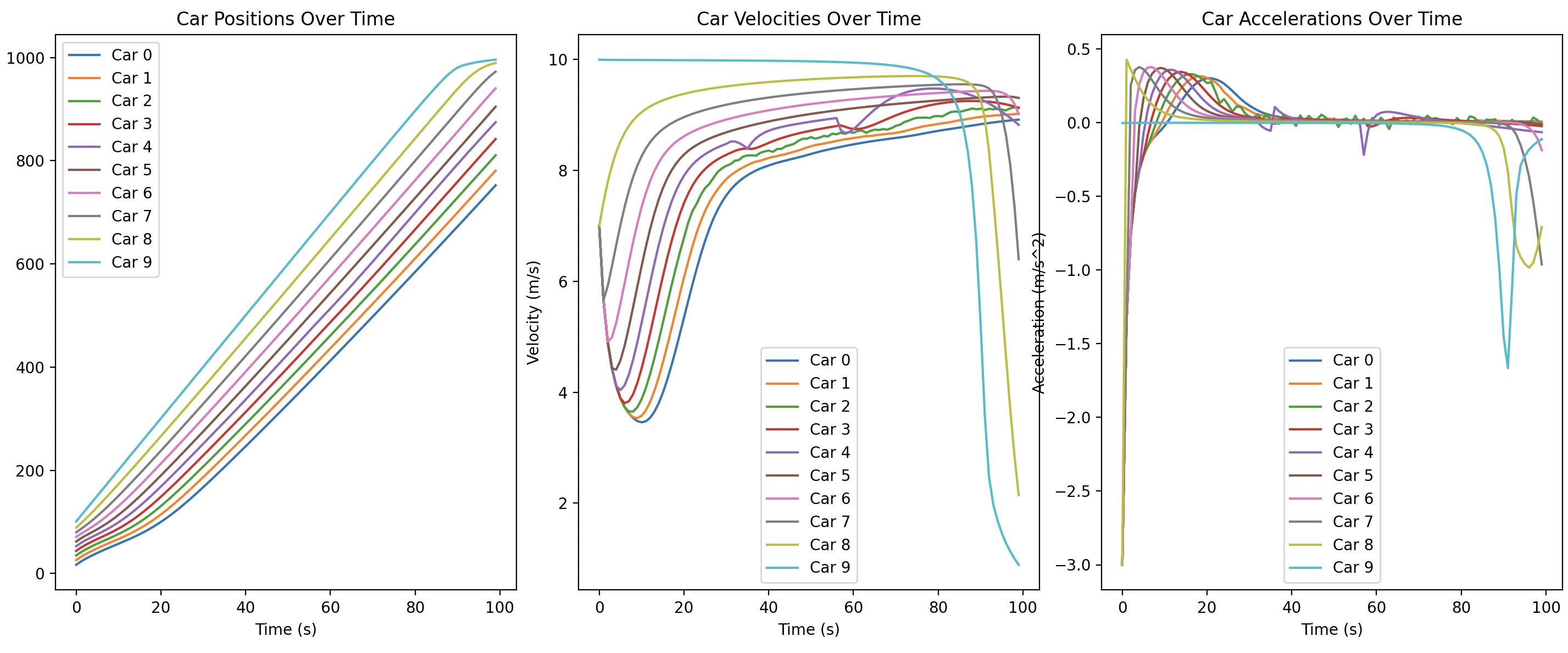}
    \caption{Car 5 experiences an exponential gap in the perceived gap }
    \label{fig:f6}
\end{figure}

Figure \ref{fig:f5} shows the impact of random fluctuations introduced to Car 2's perceived gap at $t=20$ seconds. A noise factor is added, drawn from a uniform range between -2 and 2. This random noise leads to spiky variations in both velocity and acceleration, causing instability in Car 2’s motion as it tries to adjust to the unpredictable gap. The disturbances in position, velocity, and acceleration patterns provide clear indicators of anomalies that can be detected by anomaly detection models trained on normal driving behavior. And In Figure \ref{fig:f6}, Car 4's perceived gap to the lead vehicle is reduced gradually using an exponential decay function. This causes the vehicle to frequently adjust its acceleration and deceleration, resulting in oscillations and instability as Car 4 continuously compensates for the shrinking gap. Each of these figures demonstrates the unique impact of each attack on vehicle dynamics. The disturbances in position, velocity, and acceleration patterns provide clear indicators of anomalies that can be detected by anomaly detection models trained on normal driving behavior.

\subsection{Random Forest Model Results}
The Random Forest model uses a structured input-output setup to predict the future positions of 10 autonomous vehicles. The input consists of 150 features, representing the positions of each vehicle over the past 15 seconds (10 cars × 15-time steps). This historical data provides context on each car's trajectory, which the model uses to learn typical driving patterns. The output consists of 50 target features, representing the predicted positions of these 10 vehicles over the next 5 seconds (10 cars × 5-time steps). This setup enables the model to forecast near-term vehicle positions based on recent movement trends, allowing it to identify deviations and detect anomalies when actual positions deviate significantly from the predicted values.

\begin{table}[htbp]
\centering
\caption{Random Forest Model Results}
\begin{tabular}{|l|c|}
\hline
\textbf{Metric} & \textbf{Value} \\
\hline
Mean Absolute Percentage Error (MAPE) & 0.0012 \\
Mean Absolute Error (MAE) & 5.746 \\
Mean Squared Error (MSE) & 52.37 \\
R-squared (\(R^2\)) & 0.9830 \\
95th Percentile Threshold (Anomaly) & 14.18 \\
\hline
\end{tabular}
\label{tab:rf_results}
\end{table}

\begin{figure}[htbp]
    \centering
    \includegraphics[width=1\linewidth]{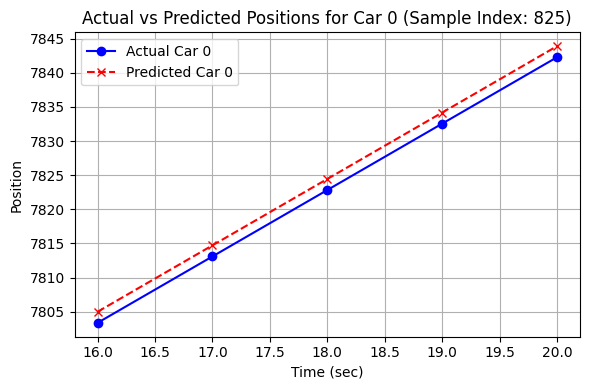}
    \caption{Actual vs predicted positions for Car 0 (Sample Test Data)}
    \label{fig:f7}
\end{figure}

The sample plot in Figure \ref{fig:f7} illustrates the comparison between actual and predicted positions for Car 0 over a 5-second interval using the Random Forest model. The close alignment between the two trajectories indicates the model's accuracy in capturing the car's movement dynamics. Minimal deviations between the actual and predicted positions highlight the model’s effectiveness in understanding vehicle behavior. This level of precision is crucial for ensuring safe and reliable navigation in autonomous vehicle systems. The consistency in predictions demonstrates the model’s robustness, even when applied to real-world trajectory data. Such accurate predictions are vital for real-time anomaly detection and proactive response in connected autonomous Vehicles.

\subsection{Stacked LSTM Model Results}
In this study, a stacked LSTM model was developed to predict future vehicle trajectories in connected autonomous vehicles (CAVs) using 15 seconds of historical position data to forecast the next 5 seconds. The model used three LSTM layers with ReLU activation to capture multi-level temporal dependencies, followed by a dense layer for output mapping. The model was trained using the Adam optimizer and MSE loss, achieving strong predictive performance on normal trajectory data. Key metrics included a low Mean Absolute Percentage Error (MAPE) of 2.04\%, indicating high accuracy, and a high R-squared (R²) value of 0.9998, showing that the model captured almost all variance in vehicle behavior. The Mean Absolute Error (MAE) was 82.43 units, reflecting close alignment with actual vehicle positions. For anomaly detection, a 95th percentile threshold of 265.63 was set based on residual errors, providing a reliable boundary to flag significant deviations as anomalies.

\begin{table}[htbp]
\centering
\caption{Stacked LSTM Model Results}
\begin{tabular}{|l|c|}
\hline
\textbf{Metric} & \textbf{Value} \\
\hline
Mean Absolute Percentage Error (MAPE) & 0.021 \\
Mean Absolute Error (MAE) & 82.425 \\
Mean Squared Error (MSE) & 13{,}636.371 \\
R-squared (\(R^2\)) & 0.9998 \\
95th Percentile Threshold (Anomaly) & 265.63 \\
\hline
\end{tabular}
\label{tab:lstm_results}
\end{table}

To detect anomalies, a 95th percentile threshold of 265.63 was established based on the residual errors observed during model training on normal trajectories. This threshold acts as a boundary, allowing the system to flag significant deviations as anomalies. By setting the threshold at this level, the model ensures that only unexpected and substantial deviations are marked, minimizing false positives and enhancing the reliability of the detection system.                     

\begin{figure}[htbp]
    \centering
    \includegraphics[width=1\linewidth]{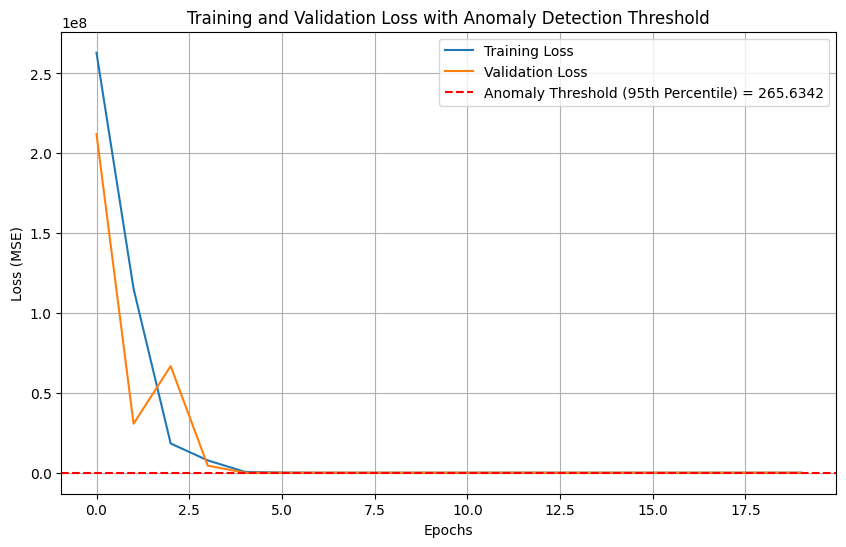}
    \caption{Training and Validation Loss using Normal Dataset}
    \label{fig:f8}
\end{figure}

\section{Conclusion}
This research develops a framework for detecting anomalies in connected autonomous vehicles (CAVs) using deep learning and machine learning. The IDM was used to create a dataset of vehicle trajectories under normal and attacked conditions for training. The Stacked LSTM model, designed to capture time-based patterns, achieved a high R-squared (R²) of 0.9998 and a MAPE of 2.04\%, showing accurate position predictions. However, its MAE of 82.43 units suggests some precision limitations, especially for small distances. A 95th percentile threshold of 265.63 was used to flag significant anomalies, though smaller disruptions may be missed. The Random Forest model showed strong performance with an R² of 0.9830, a MAPE of 0.0012, and an MAE of 5.746, indicating accurate predictions. For anomaly detection, a threshold of 14.18 effectively identified unusual behaviors while remaining sensitive to changes. Both models demonstrated strong anomaly detection, with the LSTM excelling in time-dependent patterns and the Random Forest offering precise position predictions. Future work could improve LSTM precision, refine thresholds, and explore advanced techniques like federated learning to enhance resilience against cyber threats. 

\section*{Acknowledgment}
This research was supported by the USDOT Tier-1 University Transportation Center (UTC) Transportation Cybersecurity Center for Advanced Research and Education (CYBER-CARE) (Grant No. 69A3552348332).

\bibliographystyle{unsrtnat}
\bibliography{ref}

\begin{thebibliography}{34}
\providecommand{\natexlab}[1]{#1}
\providecommand{\url}[1]{\texttt{#1}}
\expandafter\ifx\csname urlstyle\endcsname\relax
  \providecommand{\doi}[1]{doi: #1}\else
  \providecommand{\doi}{doi: \begingroup \urlstyle{rm}\Url}\fi

\bibitem[Guler et~al.(2014)Guler, Menendez, and Meier]{guler2014using}
S~Ilgin Guler, Monica Menendez, and Linus Meier.
\newblock Using connected vehicle technology to improve the efficiency of intersections.
\newblock \emph{Transportation Research Part C: Emerging Technologies}, 46:\penalty0 121--131, 2014.

\bibitem[Siegel et~al.(2017)Siegel, Erb, and Sarma]{siegel2017survey}
Joshua~E Siegel, Dylan~C Erb, and Sanjay~E Sarma.
\newblock A survey of the connected vehicle landscape—architectures, enabling technologies, applications, and development areas.
\newblock \emph{IEEE Transactions on Intelligent Transportation Systems}, 19\penalty0 (8):\penalty0 2391--2406, 2017.

\bibitem[Doecke et~al.(2015)Doecke, Grant, and Anderson]{doecke2015real}
Sam Doecke, Alex Grant, and Robert~WG Anderson.
\newblock The real-world safety potential of connected vehicle technology.
\newblock \emph{Traffic injury prevention}, 16\penalty0 (sup1):\penalty0 S31--S35, 2015.

\bibitem[Beak et~al.(2017)Beak, Head, and Feng]{beak2017adaptive}
Byungho Beak, K~Larry Head, and Yiheng Feng.
\newblock Adaptive coordination based on connected vehicle technology.
\newblock \emph{Transportation Research Record}, 2619\penalty0 (1):\penalty0 1--12, 2017.

\bibitem[Tideman and Van~Noort(2013)]{tideman2013simulation}
Martijn Tideman and Martijn Van~Noort.
\newblock A simulation tool suite for developing connected vehicle systems.
\newblock In \emph{2013 IEEE Intelligent Vehicles Symposium (IV)}, pages 713--718. IEEE, 2013.

\bibitem[Ghoul and Sayed(2021)]{ghoul2021real}
Tarek Ghoul and Tarek Sayed.
\newblock Real-time signal-vehicle coupled control: An application of connected vehicle data to improve intersection safety.
\newblock \emph{Accident Analysis \& Prevention}, 162:\penalty0 106389, 2021.

\bibitem[Salek et~al.(2022)Salek, Khan, Rahman, Deng, Islam, Khan, Chowdhury, and Shue]{salek2022review}
M~Sabbir Salek, Sakib~Mahmud Khan, Mizanur Rahman, Hsien-Wen Deng, Mhafuzul Islam, Zadid Khan, Mahsrur Chowdhury, and Mitch Shue.
\newblock A review on cybersecurity of cloud computing for supporting connected vehicle applications.
\newblock \emph{IEEE Internet of Things Journal}, 9\penalty0 (11):\penalty0 8250--8268, 2022.

\bibitem[Gao et~al.(2025)Gao, Liu, Chen, Liu, Zhang, and Sun]{gao2025exploring}
Lu~Gao, Yongxin Liu, Hongyun Chen, Dahai Liu, Yunpeng Zhang, and Jingran Sun.
\newblock Exploring traffic simulation and cybersecurity strategies using large language models.
\newblock In \emph{2025 IEEE Security and Privacy Workshops (SPW)}, pages 346--351. IEEE Computer Society, 2025.

\bibitem[Rajbahadur et~al.(2018)Rajbahadur, Malton, Walenstein, and Hassan]{rajbahadur2018survey}
Gopi~Krishnan Rajbahadur, Andrew~J Malton, Andrew Walenstein, and Ahmed~E Hassan.
\newblock A survey of anomaly detection for connected vehicle cybersecurity and safety.
\newblock In \emph{2018 IEEE Intelligent Vehicles Symposium (IV)}, pages 421--426. IEEE, 2018.

\bibitem[Eziama et~al.(2020)Eziama, Awin, Ahmed, Marina Santos-Jaimes, Pelumi, and Corral-De-Witt]{eziama2020detection}
Elvin Eziama, Faroq Awin, Sabbir Ahmed, Luz Marina Santos-Jaimes, Akinyemi Pelumi, and Danilo Corral-De-Witt.
\newblock Detection and identification of malicious cyber-attacks in connected and automated vehicles’ real-time sensors.
\newblock \emph{Applied Sciences}, 10\penalty0 (21):\penalty0 7833, 2020.

\bibitem[Wang et~al.(2020)Wang, Masoud, and Khojandi]{wang2020anomaly}
Yiyang Wang, Neda Masoud, and Anahita Khojandi.
\newblock Anomaly detection in connected and automated vehicles using an augmented state formulation.
\newblock In \emph{2020 Forum on Integrated and Sustainable Transportation Systems (FISTS)}, pages 156--161. IEEE, 2020.

\bibitem[Khanmohammadi and Azmi(2024)]{khanmohammadi2024time}
Fatemeh Khanmohammadi and Reza Azmi.
\newblock Time-series anomaly detection in automated vehicles using d-cnn-lstm autoencoder.
\newblock \emph{IEEE Transactions on Intelligent Transportation Systems}, 2024.

\bibitem[Mansourian et~al.(2023)Mansourian, Zhang, Jaekel, and Kneppers]{mansourian2023deep}
Pegah Mansourian, Ning Zhang, Arunita Jaekel, and Marc Kneppers.
\newblock Deep learning-based anomaly detection for connected autonomous vehicles using spatiotemporal information.
\newblock \emph{IEEE Transactions on Intelligent Transportation Systems}, 24\penalty0 (12):\penalty0 16006--16017, 2023.

\bibitem[Wang et~al.(2024)Wang, Li, and Khattak]{wang2024anomaly}
Boyu Wang, Wan Li, and Zulqarnain~H Khattak.
\newblock Anomaly detection in connected and autonomous vehicle trajectories using lstm autoencoder and gaussian mixture model.
\newblock \emph{Electronics}, 13\penalty0 (7):\penalty0 1251, 2024.

\bibitem[Gay et~al.(2015)Gay, Kniss, John, et~al.]{gay2015safety}
Kevin Gay, Valerie Kniss, A~John, et~al.
\newblock Safety pilot model deployment: lessons learned and recommendations for future connected vehicle activities.
\newblock Technical report, United States. Department of Transportation. Intelligent Transportation~…, 2015.

\bibitem[Kang et~al.(2021)Kang, Kwak, Lee, Lee, Lee, and Kim]{kang2021car}
Hyunjae Kang, Byung~Il Kwak, Young~Hun Lee, Haneol Lee, Hwejae Lee, and Huy~Kang Kim.
\newblock Car hacking: Attack defense challenge 2020 dataset.
\newblock \emph{(No Title)}, 2021.

\bibitem[Tiernan et~al.(2017)Tiernan, Richardson, Azeredo, Najm, Lochrane, John, et~al.]{tiernan2017test}
Tim Tiernan, Nicholas Richardson, Philip Azeredo, Wassim~G Najm, Taylor Lochrane, A~John, et~al.
\newblock Test and evaluation of vehicle platooning proof-of-concept based on cooperative adaptive cruise control.
\newblock Technical report, John A. Volpe National Transportation Systems Center (US), 2017.

\bibitem[Wang et~al.(2019)Wang, Mavromatis, Tassi, Santos-Rodriguez, and Piechocki]{wang2019location}
Xiaoyang Wang, Ioannis Mavromatis, Andrea Tassi, Raul Santos-Rodriguez, and Robert~J Piechocki.
\newblock Location anomalies detection for connected and autonomous vehicles.
\newblock In \emph{2019 IEEE 2nd Connected and Automated Vehicles Symposium (CAVS)}, pages 1--5. IEEE, 2019.

\bibitem[Devika et~al.(2024)Devika, Shrivastava, Narang, Alladi, and Yu]{devika2024vadgan}
S~Devika, Rishi~Rakesh Shrivastava, Pratik Narang, Tejasvi Alladi, and F~Richard Yu.
\newblock Vadgan: An unsupervised gan framework for enhanced anomaly detection in connected and autonomous vehicles.
\newblock \emph{IEEE Transactions on Vehicular Technology}, 2024.

\bibitem[Dong et~al.(2020)Dong, Wang, Ni, Liu, and Chen]{dong2020impact}
Changyin Dong, Hao Wang, Daiheng Ni, Yongfei Liu, and Quan Chen.
\newblock Impact evaluation of cyber-attacks on traffic flow of connected and automated vehicles.
\newblock \emph{IEEE Access}, 8:\penalty0 86824--86835, 2020.

\bibitem[Wang et~al.(2021{\natexlab{a}})Wang, Wu, Lin, Boddupalli, Chang, Lin, Shih, and Ray]{wang2021deep}
Sheng-Li Wang, Sing-Yao Wu, Ching-Chu Lin, Srivalli Boddupalli, Po-Jui Chang, Chung-Wei Lin, Chi-Sheng Shih, and Sandip Ray.
\newblock Deep-learning-based intrusion detection for autonomous vehicle-following systems.
\newblock In \emph{2021 IEEE International Intelligent Transportation Systems Conference (ITSC)}, pages 865--872. IEEE, 2021{\natexlab{a}}.

\bibitem[Malhotra et~al.(2015)Malhotra, Vig, Shroff, Agarwal, et~al.]{malhotra2015long}
Pankaj Malhotra, Lovekesh Vig, Gautam Shroff, Puneet Agarwal, et~al.
\newblock Long short term memory networks for anomaly detection in time series.
\newblock In \emph{Proceedings}, volume~89, page~94, 2015.

\bibitem[Sun et~al.(2021)Sun, Yu, and Zhang]{sun2021survey}
Xiaoqiang Sun, F~Richard Yu, and Peng Zhang.
\newblock A survey on cyber-security of connected and autonomous vehicles (cavs).
\newblock \emph{IEEE Transactions on Intelligent Transportation Systems}, 23\penalty0 (7):\penalty0 6240--6259, 2021.

\bibitem[Cheng et~al.(2019)Cheng, Chen, De~Vos, Lai, and Witlox]{cheng2019applying}
Long Cheng, Xuewu Chen, Jonas De~Vos, Xinjun Lai, and Frank Witlox.
\newblock Applying a random forest method approach to model travel mode choice behavior.
\newblock \emph{Travel behaviour and society}, 14:\penalty0 1--10, 2019.

\bibitem[Manzoor and Morgan(2018)]{manzoor2018vehicle}
Muhammad~Asif Manzoor and Yasser Morgan.
\newblock Vehicle make and model recognition using random forest classification for intelligent transportation systems.
\newblock In \emph{2018 IEEE 8th Annual Computing and Communication Workshop and Conference (CCWC)}, pages 148--154. IEEE, 2018.

\bibitem[Gao et~al.(2021)Gao, Lu, and Ren]{gao2021deep}
Lu~Gao, Pan Lu, and Yihao Ren.
\newblock A deep learning approach for imbalanced crash data in predicting highway-rail grade crossings accidents.
\newblock \emph{Reliability Engineering \& System Safety}, 216:\penalty0 108019, 2021.

\bibitem[Parosh~Yamarthi et~al.(2025)Parosh~Yamarthi, Raman, and Parvin]{parosh2025united}
Dominic Parosh~Yamarthi, Haripriya Raman, and Shamsad Parvin.
\newblock United states road accident prediction using random forest predictor.
\newblock \emph{arXiv e-prints}, pages arXiv--2505, 2025.

\bibitem[Elsadig et~al.(2025)Elsadig, Altigani, Mohamed, Mohamed, Kannan, Bashir, and Adiel]{elsadig2025connected}
Muawia~A Elsadig, Abdelrahman Altigani, Yasir Mohamed, Abdul~Hakim Mohamed, Akbar Kannan, Mohamed Bashir, and Mousab~AE Adiel.
\newblock Connected vehicles security: A lightweight machine learning model to detect vanet attacks.
\newblock \emph{World Electric Vehicle Journal}, 16\penalty0 (6):\penalty0 324, 2025.

\bibitem[Gao et~al.(2022)Gao, Yu, and Lu]{gao2022missing}
Lu~Gao, Ke~Yu, and Pan Lu.
\newblock Missing pavement performance data imputation using graph neural networks.
\newblock \emph{Transportation research record}, 2676\penalty0 (12):\penalty0 409--419, 2022.

\bibitem[Siami-Namini et~al.(2019)Siami-Namini, Tavakoli, and Namin]{siami2019performance}
Sima Siami-Namini, Neda Tavakoli, and Akbar~Siami Namin.
\newblock The performance of lstm and bilstm in forecasting time series.
\newblock In \emph{2019 IEEE International conference on big data (Big Data)}, pages 3285--3292. IEEE, 2019.

\bibitem[Kumar et~al.(2021)Kumar, Damaraju, Kumar, Kumari, and Chen]{kumar2021lstm}
Sachin Kumar, Agam Damaraju, Aditya Kumar, Saru Kumari, and Chien-Ming Chen.
\newblock Lstm network for transportation mode detection.
\newblock \emph{Journal of Internet Technology}, 22\penalty0 (4):\penalty0 891--902, 2021.

\bibitem[Gao et~al.(2023)Gao, Han, and Chen]{gao2023deep}
Lu~Gao, Zhe Han, and Yunshen Chen.
\newblock Deep learning--based pavement performance modeling using multiple distress indicators and road work history.
\newblock \emph{Journal of Transportation Engineering, Part B: Pavements}, 149\penalty0 (1):\penalty0 04022061, 2023.

\bibitem[Wang et~al.(2021{\natexlab{b}})Wang, Li, and Yue]{wang2021inclstm}
Huiju Wang, Mengxuan Li, and Xiao Yue.
\newblock Inclstm: incremental ensemble lstm model towards time series data.
\newblock \emph{Computers \& Electrical Engineering}, 92:\penalty0 107156, 2021{\natexlab{b}}.

\bibitem[Silwal et~al.(2024)Silwal, Gao, Zhang, Senouci, and Mo]{silwal2024assessing}
Saurav Silwal, Lu~Gao, Yunpeng Zhang, Ahmed Senouci, and Yi-Lung Mo.
\newblock Assessing cybersecurity risks and traffic impact in connected autonomous vehicles.
\newblock In \emph{International Conference on Transportation and Development 2024}, pages 652--662, 2024.

\end{thebibliography}

\end{document}